\documentclass[suppldata]{interact}

\usepackage{epstopdf}
\usepackage[caption=false]{subfig}

\usepackage[longnamesfirst,sort]{natbib}
\bibpunct[, ]{(}{)}{;}{a}{,}{,}
\renewcommand\bibfont{\fontsize{10}{12}\selectfont}

\theoremstyle{plain}

\theoremstyle{definition}

\theoremstyle{remark}

\usepackage[colorinlistoftodos]{todonotes}
\usepackage{pdfcomment}
\usepackage{subcaption}
\usepackage{graphicx}
\usepackage{float}

\begin{document}

\title{Machine Learning for Detection and Severity Estimation of Sweetpotato Weevil Damage in Field and Lab Conditions}

\author{
\name{Doreen M. Chelangat\textsuperscript{a}, Sudi Murindanyi\textsuperscript{b}, Bruce Mugizi\textsuperscript{c}, Paul Musana\textsuperscript{a}, Benard Yada\textsuperscript{a}, Milton A. Otema\textsuperscript{a}, Florence Osaru\textsuperscript{a}, Andrew Katumba\textsuperscript{c}, Joyce Nakatumba-Nabende\textsuperscript{b}}
\vspace{0.2cm}
\affil{\textsuperscript{a}National Crops Resources Research Institute (NaCRRI), Uganda\\
\textsuperscript{b}Department of Computer Science, Makerere University, Uganda\\
\textsuperscript{c}Department of Electrical and Computer Engineering, Makerere University, Uganda
}}
\maketitle
\begin{abstract}
    Sweetpotato weevils (\textit{Cylas} spp.) are considered among the most destructive pests impacting sweetpotato production, particularly in sub-Saharan Africa. Traditional methods for assessing weevil damage, predominantly relying on manual scoring, are labour-intensive, subjective, and often yield inconsistent results. These challenges significantly hinder breeding programs aimed at developing resilient sweetpotato varieties. This study introduces a computer vision-based approach for the automated evaluation of weevil damage in both field and laboratory contexts. In the field settings, we collected data to train classification models to predict root-damage severity levels, achieving a test accuracy of 71.43\%. Additionally, we established a laboratory dataset and designed an object detection pipeline employing YOLO12, a leading real-time detection model. This methodology incorporated a two-stage laboratory pipeline that combined root segmentation with a tiling strategy to improve the detectability of small objects. The resulting model demonstrated a mean average precision of 77.7\% in identifying minute weevil feeding holes. Our findings indicate that computer vision technologies can provide efficient, objective, and scalable assessment tools that align seamlessly with contemporary breeding workflows. These advancements represent a significant improvement in enhancing phenotyping efficiency within sweetpotato breeding programs and play a crucial role in mitigating the detrimental effects of weevils on food security.
\end{abstract}

\section{Introduction}
Sweetpotatoes are a vital staple crop in sub-Saharan Africa, providing both food security and income, particularly in Uganda. Sweet potatoes rank among the top three root crops often grown by smallholder farmers, and have a strong presence in urban and rural markets \citep{FAO, San15, Mud19, Liu14}. They are a crucial crop for economic stability in many regions, providing essential nutrients and serving as a vital source of income for local communities. Its low input requirements and drought resilience make it a suitable crop for small-scale farming, and it has the potential to improve rural well-being \citep{You20, Mot15}.

However, the African sweetpotato weevil (\textit{Cylas puncticollis/brunneus}) species poses a severe threat. These pests burrow into roots, sometimes causing yield losses up to 100\%, reducing root quality and causing damage that makes the roots unfit for human consumption \citep{Oko16, Muy12, CIPbrunneus, florencemorphology}. To reduce this challenge, agricultural experts assess root damage to identify weevil-resistant genotypes - genetic variants within a species that exhibit specific inherited traits \citep{Mug22, Any13, Muy12}. This assessment process involves assigning weevil severity scores based on standardised evaluation criteria, enabling researchers to distinguish between susceptible and resistant varieties \citep{Mug22}. The identification of resistant genotypes is crucial for developing improved varieties that farmers can adopt to reduce crop losses and ensure food security \citep{Yad23, florencemorphology, Fit17}.

Currently, phenotyping sweetpotato varieties for weevil resistance relies entirely on manual assessment methods. In field evaluations, experts conduct visual scoring using a standardised 1-9 scale to assess damage severity across plots \citep{gruneberg2019sweetpotato, Aka18, florencemorphology}, while laboratory (lab) assessments involve counting individual feeding holes following controlled weevil infestation protocols \citep{Mug22}. These approaches are subjective, labour-intensive, and time-consuming, requiring extensive expert involvement and experimental protocols \citep{Ruk13, florencemorphology, Mug22}. Furthermore, the gaps created by these manual assessment methods can significantly delay breeding programs targeting weevil resistance, potentially hindering the timely development of improved varieties needed to meet growing agricultural demands \citep{Mak24, CIPBiotech}.

Advancements in machine learning and deep learning offer significant potential to overcome these limitations. Image-based machine learning approaches can automate both detection and severity scoring of weevil infestations, substantially improving assessment reliability and throughput while reducing dependence on expert labour \citep{florencemorphology, katumba2024leveraging}. This addresses the subjectivity of manual scoring by providing more consistent, controlled, and repeatable assessments.
 
Our study introduces imaging pipelines designed for both field and laboratory contexts: field assessments involve capturing images of harvested roots from experimental plots alongside expert visual scoring (1-9 scale), while laboratory assessments focus on individual roots exposed to controlled weevil infestations, with subsequent imaging to document feeding damage patterns and enable precise feeding hole counting. Our approach applied convolutional neural networks (CNNs) for plot-level severity scoring and estimation in the field, while laboratory data analysis utilised a two-stage detection process. First segments the root region of interest to isolate relevant areas, then applies a tiling strategy using the Slicing Aided Hyper Inference (SAHI) framework \citep{Aky22} to detect tiny weevil feeding holes that would otherwise be missed in full-resolution images.

A secondary emphasis of this research was ensuring real-world applicability by developing lightweight machine learning models that can be deployed on resource-constrained edge devices, such as smartphones or portable tools, accessible to farmers and field technicians. This required striking a careful trade-off between model complexity (accuracy) and deployment efficiency (speed and size). To this end, we focused on fine-tuning models that offer faster inference times, improved efficiency, and reduced computational requirements.


This paper aimed to:
\begin{itemize}
    \item Collect and curate the dataset on field and lab datasets.
    \item Develop a pipeline for estimating weevil damage scores from field plot images.
    \item Develop a segmentation and detection pipeline for quantifying feeding damage in laboratory-infested roots.
    \item Establish a foundation for providing a dataset for weevil damage assessment in future work.
\end{itemize}

By addressing both field and lab phenotyping challenges, we seek to provide a scalable, objective, and accessible toolset that can significantly reduce subjectivity,  manual effort, speed up breeding for weevil resistance, and improve sweetpotato value chains in Uganda and across Africa.

The remainder of this paper is organised as follows: Section 2 reviews relevant literature. Section 3 describes the materials and methods used in the field and laboratory assessments. Section 4 details the model development process for both classification and detection tasks. Results are presented in Section 5, followed by discussion in Section 6 and conclusions in Section 7.

\section{Background}
Traditional sweetpotato phenotyping has relied heavily on manual assessment methods that are both time-consuming and subjective. Recent advances in machine learning have shown promise in automating sweetpotato root analysis, with researchers successfully deploying models to analyse sensory attributes such as fresh colour and mealiness in breeding programs at the International Potato Centre in Uganda \citep{NAKATUMBANABENDE2023100291}. Convolutional neural networks, particularly the InceptionResNetV2 architecture, have demonstrated effectiveness in qualitative improvement assessments of sweetpotato roots, reducing subjectivity in analysis and phenotyping time \citep{fernandes2023cnn}. However, these approaches have primarily focused on quality assessment rather than evaluating pest damage.

High-throughput phenotyping methods have been developed for the characterisation of sweetpotato virus resistance, utilising remotely sensed data to accurately predict resistance categories based on yield loss and virus load \citep{kreuze2024virus}. More recently, deep learning approaches have been applied to identify sweetpotato virus disease-infected leaves from field images, addressing the costly and labour-intensive nature of current diagnostic methods \citep{ding2024virus}. Despite these advances, assessment of physical pest damage, particularly from weevil infestations, remains largely unexplored in the literature.

The broad application of computer vision and deep learning in agricultural disease detection has grown rapidly, with researchers leveraging deep learning's superiority over traditional digital image processing methods for plant disease and pest identification \citep{upadhyay2025dlreview, liu2021plantpests}.
Modern approaches utilise machine learning with data from multiple sensor types to build models for detection, prediction, and assessment, addressing the serious impact of crop diseases on agricultural productivity and global food security \citep{ouhami2021survey, meraj2023phenotyping}.

Disease severity assessment using convolutional neural networks has gained particular attention, as determining not only disease type but also severity is crucial for effective monitoring and control of plant production processes \citep{stavness2023editorial}. Various imaging sensors, including RGB, multispectral, thermal, and hyperspectral, have been employed, with thermal and hyperspectral sensors proving most effective for early-stage pathogen detection, while RGB sensors remain suitable for assessing the severity of later-stage infections. Image-based crop disease detection using machine learning \citep{dolatabadian2025disease}. The urgency of this research is underscored by the fact that major food crops can suffer losses of 10\% to 40\% due to plant viruses, necessitating frequent and efficient disease inspection methods \citep{wen2022pestyolo}.

\subsection{Object Detection for Small Agricultural Objects}
Detecting small objects in agricultural settings presents unique challenges that have driven specialised model development. Pest-YOLO models have been specifically designed for large-scale, multi-class, dense, and tiny pest detection and counting, utilising YOLOv4 as a benchmark model with focal loss integration to improve performance on small objects \citep{encordyolo}. Agricultural pest detection using YOLOv5 frameworks has revealed limitations in extracting key features from pest images containing large background noise and dense object distributions \citep{datacampyolo, hakim2025yolopest}.

Recent improvements in pest detection have focused on efficient channel attention mechanisms and transformer encoders, notably enhancing representation of small agricultural pests and their feature expressiveness for real-time identification \citep{BADGUJAR2024109090}.
Optimised YOLO variants have been developed for small insect detection, featuring high speed and low memory requirements suitable for continuous monitoring and rapid decision-making in agricultural fields \citep{WANG2025110085}. The growing importance of object detection in digital farming has been recognised as essential for precision agriculture applications \citep{zhang2022agripest}.

\bigskip

Despite significant advances, several challenges persist in agricultural phenotyping and pest detection systems. The genetic behaviour of plant traits varies significantly across different growing environments, making it difficult to create reliable phenotyping systems \citep{DING2024120338}. Current computer vision systems for crop disease monitoring face challenges in optimising resource allocation and boosting productivity while maintaining detection accuracy \citep{shi2023severity}.
Traditional agricultural monitoring systems require predetermined steps involving multiple sensors and complex data collection procedures, often limiting their practical deployment \citep{anand2024disease}. The integration of multiple data sources and the development of assessment frameworks capable of handling both classification and detection tasks simultaneously remain active areas of research. Furthermore, the specific challenge of assessing physical damage from storage pests, such as sweetpotato weevils, which requires both severity classification and precise damage quantification, represents a significant gap in the current agricultural phenotyping literature.

\section{Materials and Methods}
This section is divided into two parts: field assessment and lab assessment, each detailing the data collection protocols, experimental setup, imaging procedures and preprocessing techniques applied.

\subsection{Field Assessment}

\subsubsection{Site Description}

Field phenotyping was conducted at four distinct agricultural sites in Uganda—NaSARRI, Rwabitaba, Abi-ZARDI, and Bu-ZARDI—representing diverse agroecological conditions across the country's major sweetpotato production zones \citep{florencemorphology}. Multiple breeding trials, including Preliminary Yield Trials (PYT), Advanced Yield Trials (AYT), Seed Trials (ST), Distinctness, Uniformity and Stability trials (DUS), and Genotype by Environment trials (GG), were conducted across two cropping seasons \citep{GrunebergEtAl2019, Kondwakwenda2022}. Video recordings and image captures were collected during the first and second seasons, respectively. Environmental conditions, such as rainfall patterns and planting schedules, were typical of sweetpotato production zones in Uganda. 

\subsubsection{Root Collection and Scoring}

\begin{figure}[h!]
    \centering
    
    \subfloat[Severity score: 1]{%
      \resizebox*{6.5cm}{!}{\includegraphics{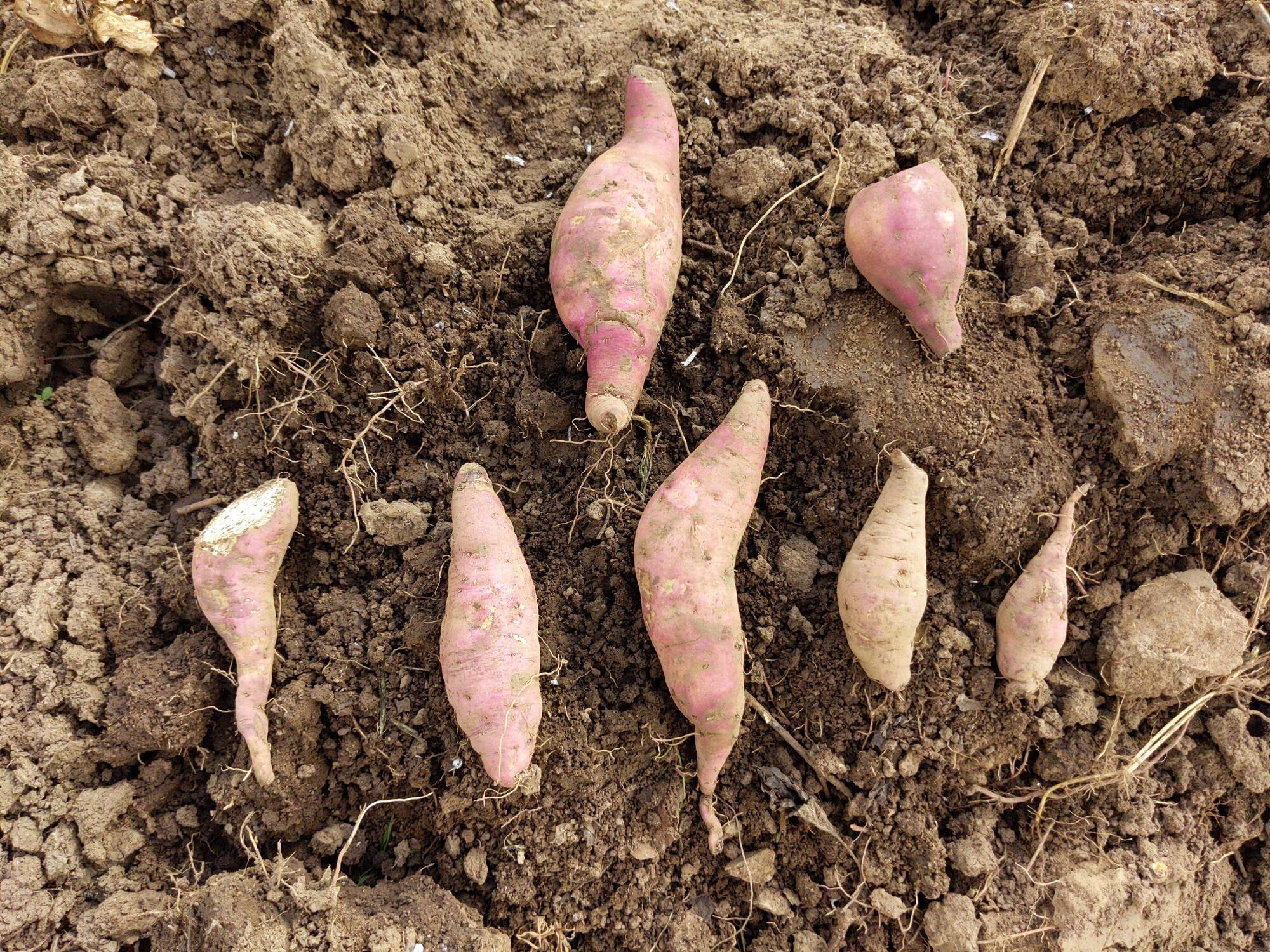}}} \hspace{5pt}
    \subfloat[Severity score: 3]{%
      \resizebox*{6.5cm}{!}{\includegraphics{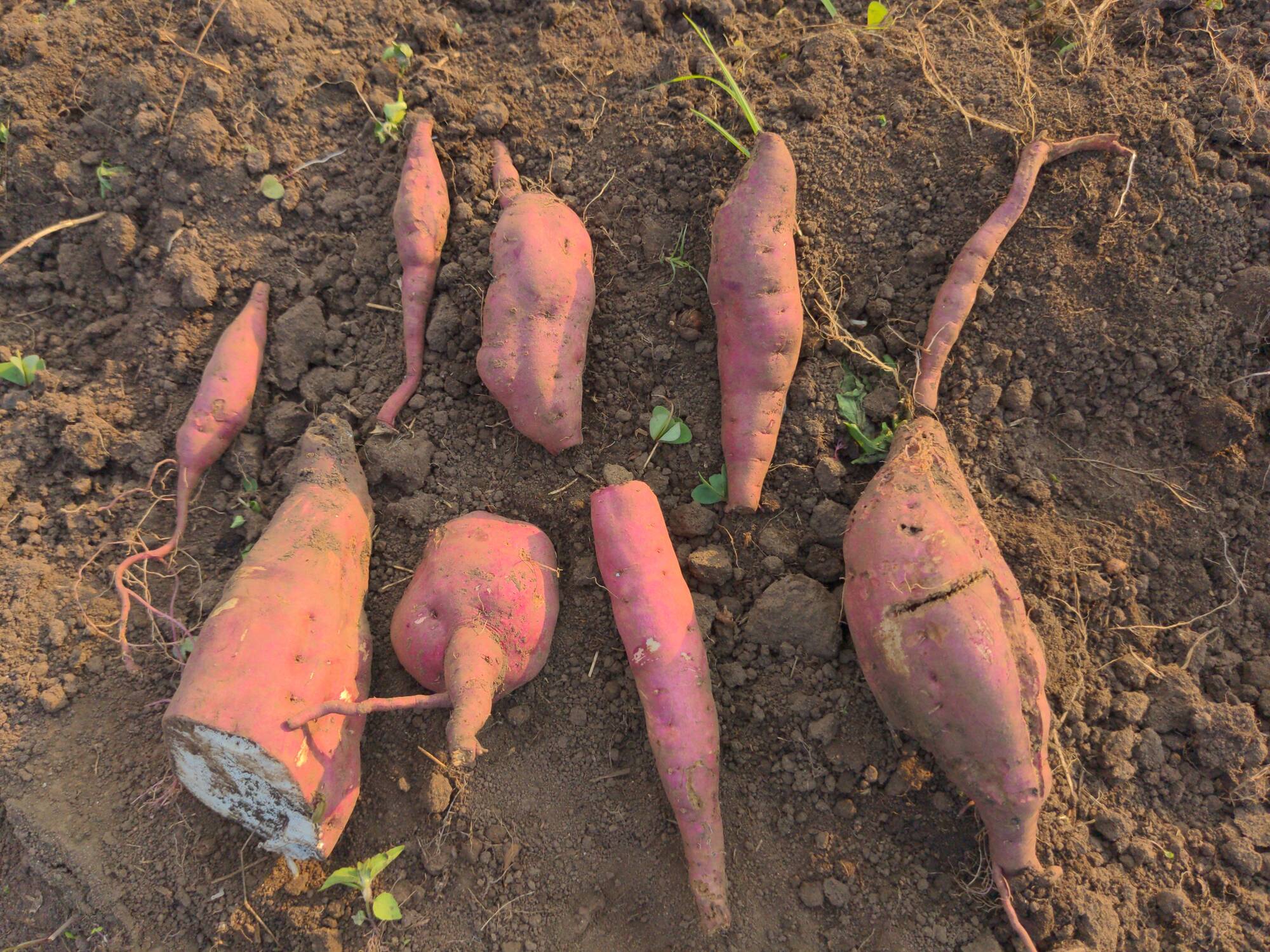}}} \hspace{5pt}
    \subfloat[Severity score: 5]{%
      \resizebox*{6.5cm}{!}{\includegraphics{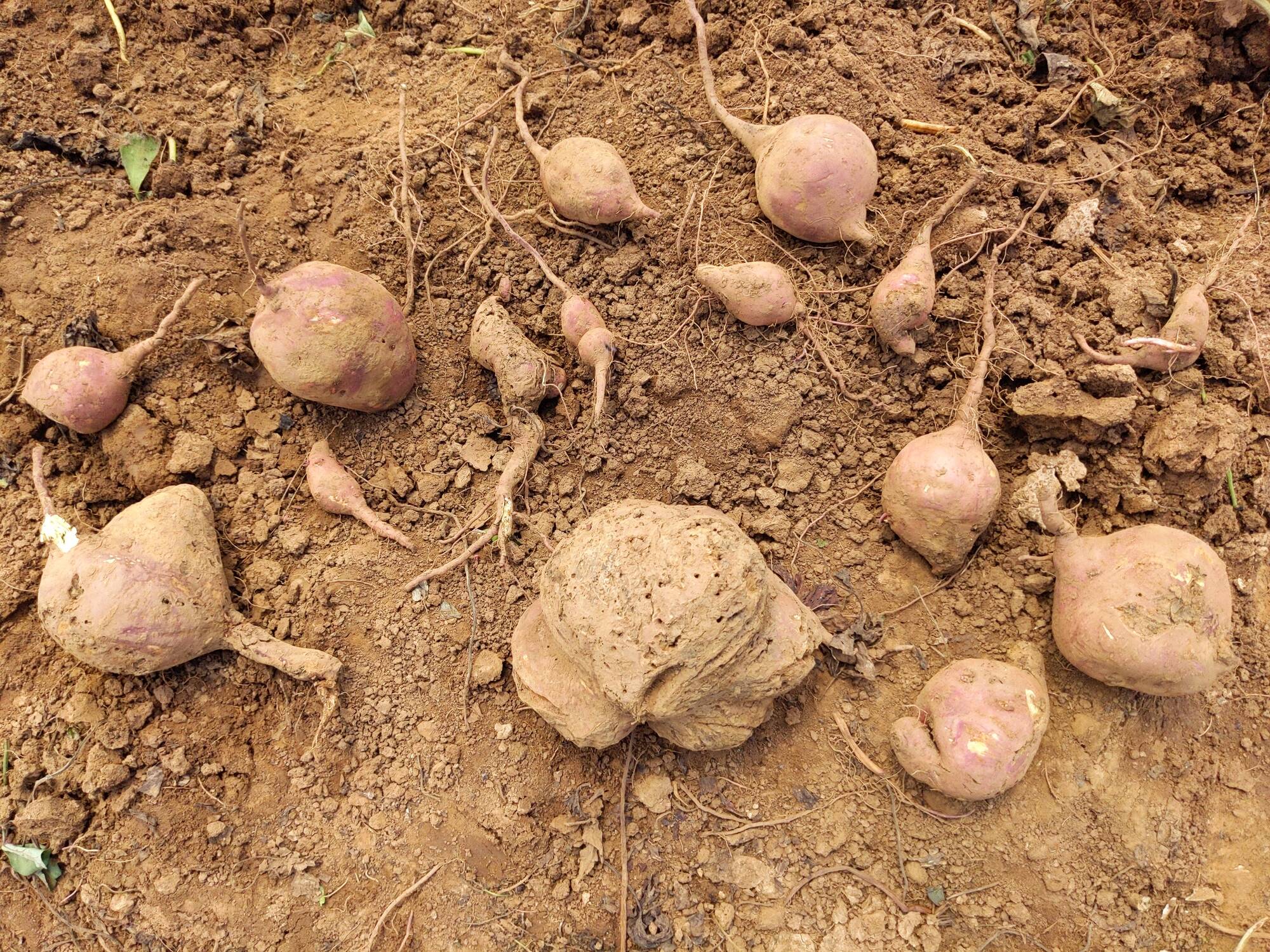}}} \hspace{5pt}
    \subfloat[Severity score: 7]{%
      \resizebox*{6.5cm}{!}{\includegraphics{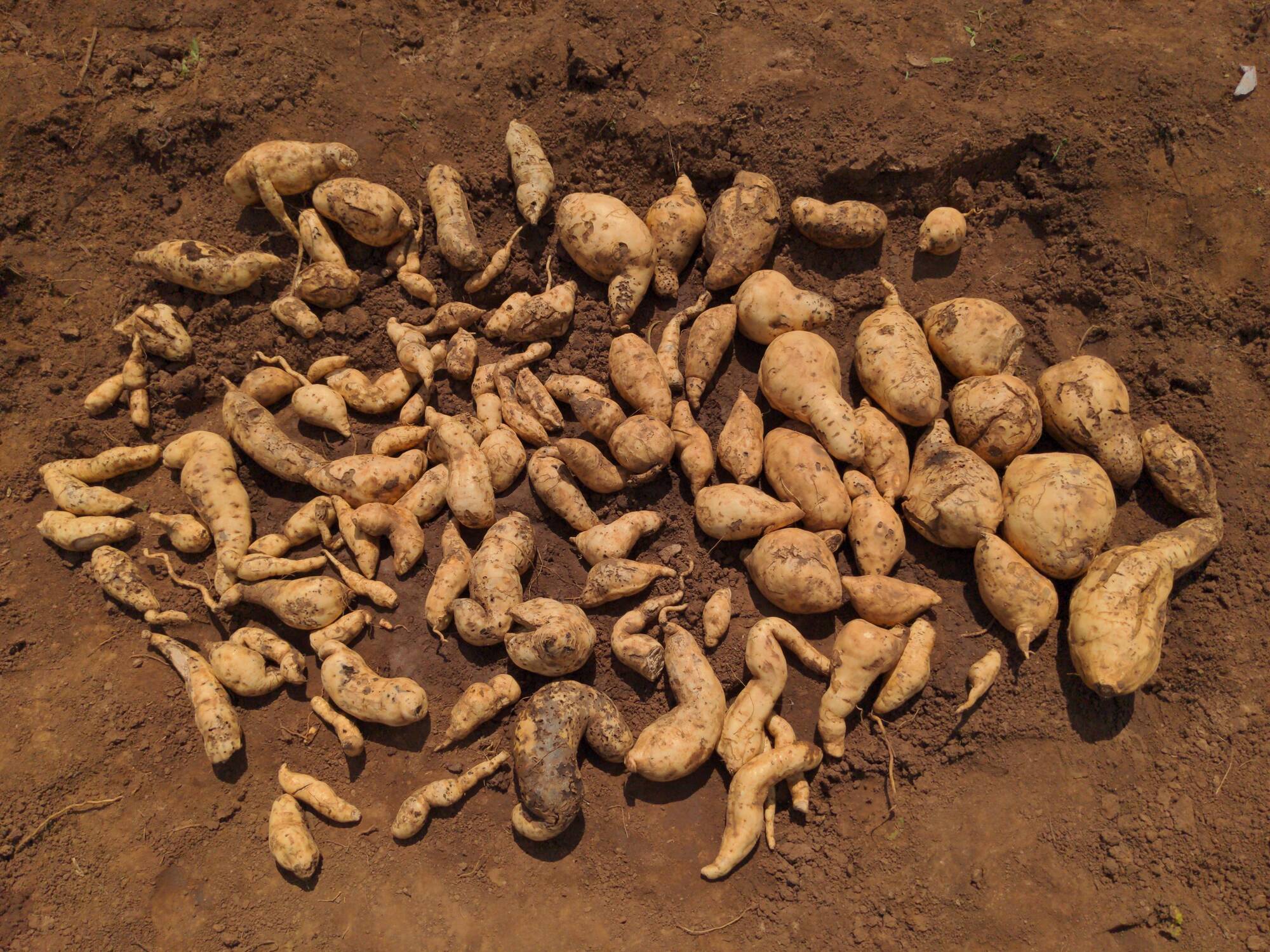}}} \hspace{5pt}
    \subfloat[Severity score: 9]{%
      \resizebox*{8cm}{!}{\includegraphics{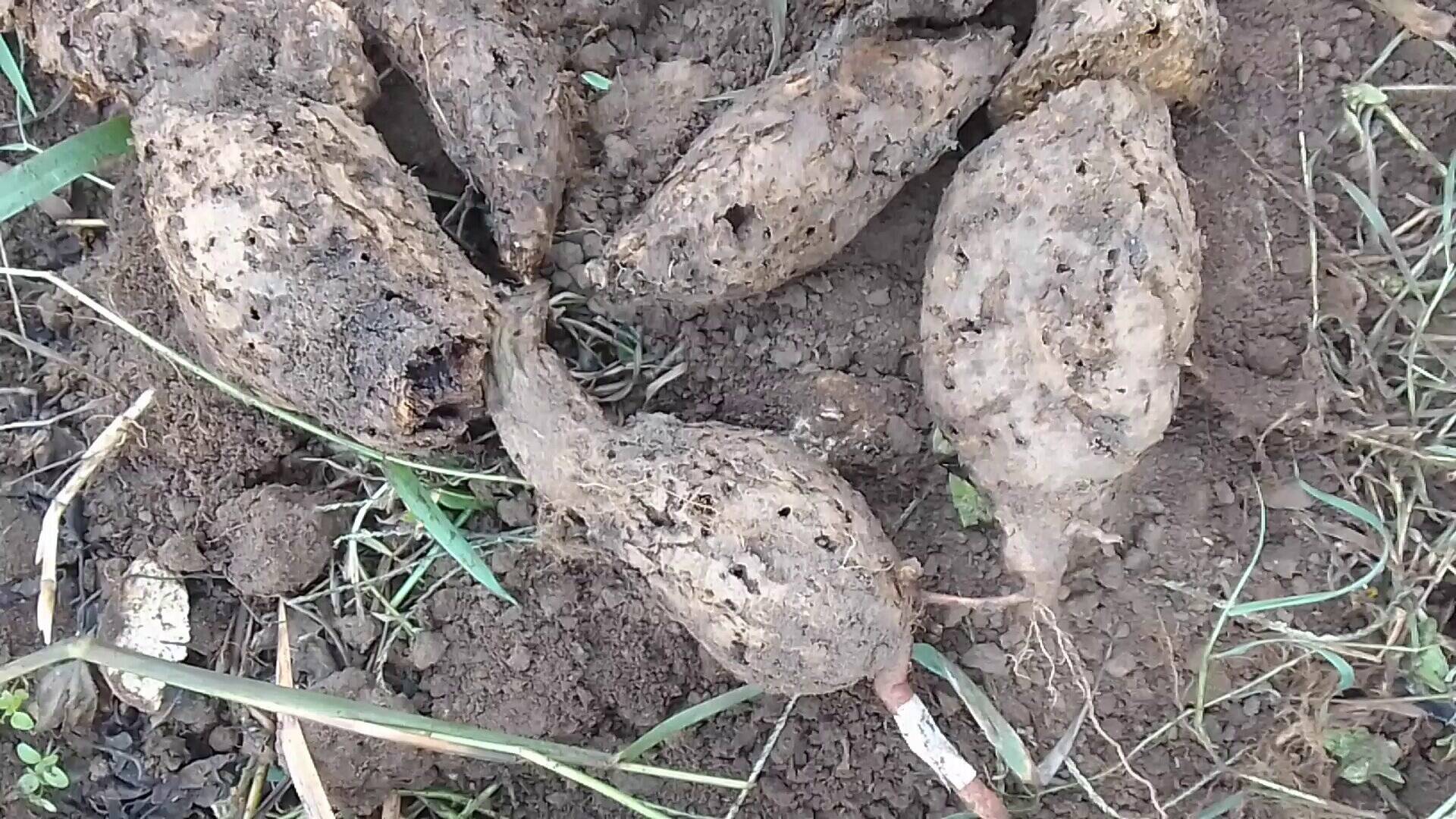}}}
    
    \caption{Example field plots illustrating the adapted 5-point weevil damage severity scale. Severity scores (1, 3, 5, 7, 9) are assigned based on the proportion of damaged roots relative to the total number of roots in each plot.}
    \label{fig:severity-scores-grid}
\end{figure}

During the harvest, vines were uprooted, and storage roots were excavated, cleaned, and arranged on cleared ground to enhance visibility of any damage. After consulting with experts, we decided to adapt the standard 1-9 scoring scale \citep{gruneberg2019sweetpotato} to utilise only scores of 1, 3, 5, 7, and 9. This adaptation was made for several reasons. First, achieving a balanced representation of plots across all scores would be challenging and could introduce bias early in the assessment process. Second, experts indicated that obtaining plots for some intermediate scores would pose difficulties. Third, there are minimal differences between closely related scores, which could complicate accurate scoring. Taking these considerations into account, we chose to adopt a 5-score scale, with experts assigning scores according to this modified framework. Table \ref{tab:field-damage-scale} presents the description of every class, and figure \ref{fig:severity-scores-grid} illustrates sample images of the plots along with their assigned severity scores.
 
\begin{table}[h]
    \centering
    \caption{The adapted 5-point damage severity classification scale used for visual assessment of weevil damage in field plots. Adapted from \citep{gruneberg2019sweetpotato}.}
    \label{tab:field-damage-scale}
    \begin{tabular}{|c|l|}
        \hline
        \textbf{Class} & \textbf{Description} \\ \hline
        1             & No visible damage \\ \hline
        3             & $<$ 5\% of roots in a plot show damage \\ \hline
        5             & 16–33\% damaged \\ \hline
        7             & 67–99\% damaged \\ \hline
        9             & All roots damaged \\ \hline
    \end{tabular}
\end{table}

\subsubsection{Image and Video Acquisition}

\begin{figure}[h!]
    \centering
    
    \subfloat[Top view: Capturing the flat upper side of the roots.]{%
      \resizebox*{6.5cm}{!}{\includegraphics{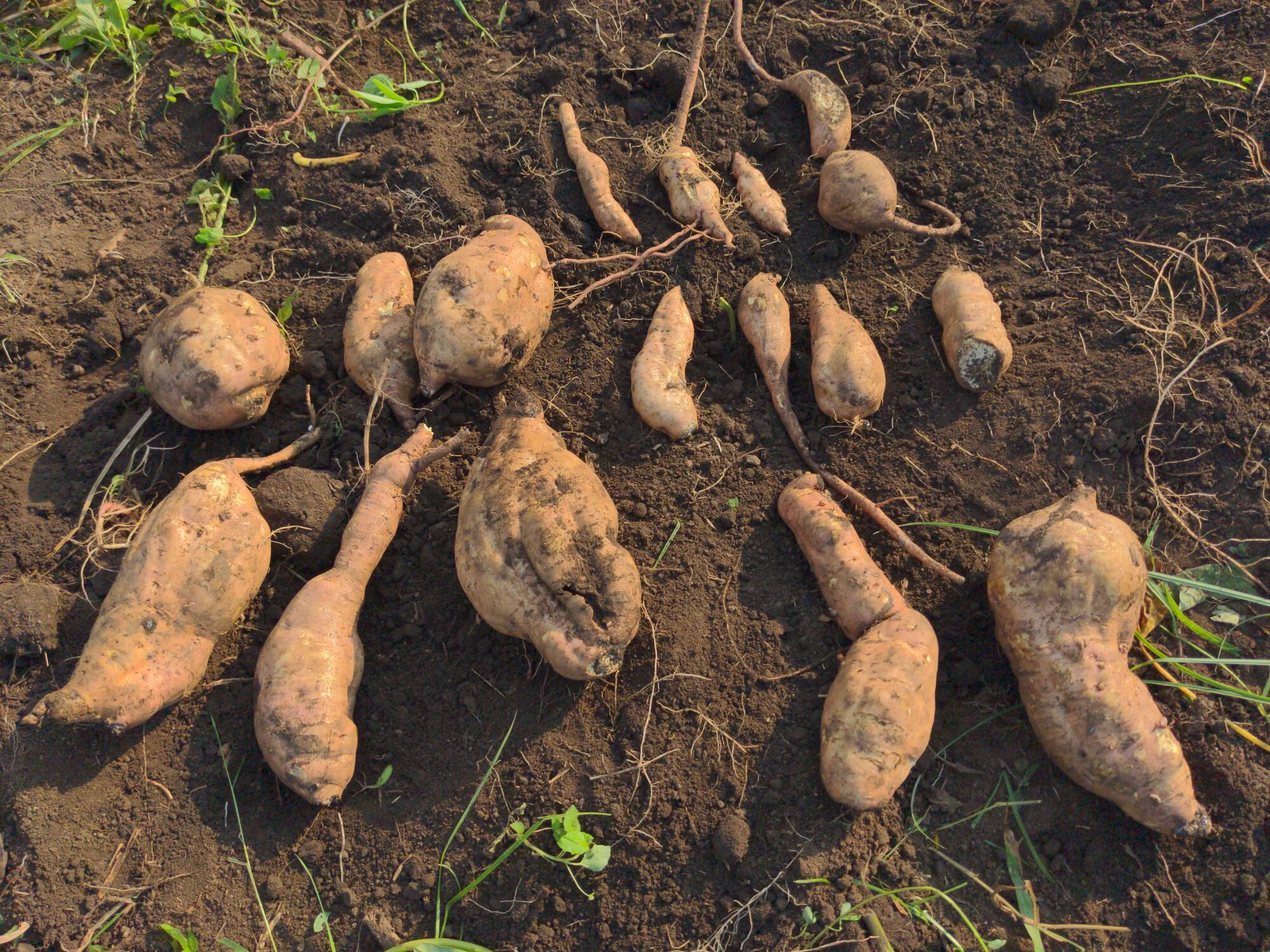}}} \hspace{10pt}
    \subfloat[Bottom view: Underside of the same roots after rotation.]{%
      \resizebox*{6.5cm}{!}{\includegraphics{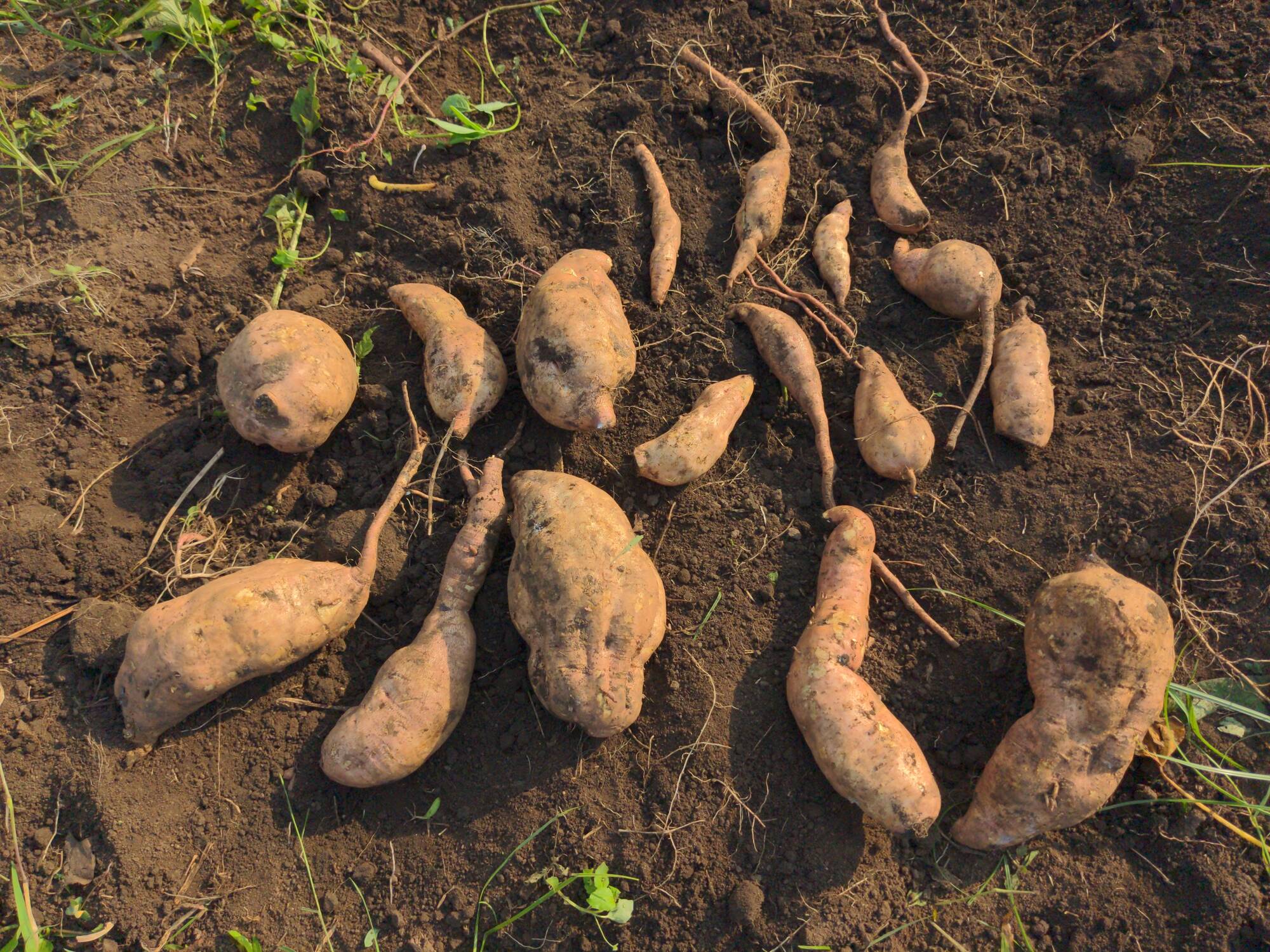}}}
    
    \caption{Orthogonal views of storage roots captured at the plot level. The left image shows the top view, taken as the roots were initially arranged. The right image shows the bottom view, captured after rotating the roots to reveal the underside.}
    \label{fig:orthogonal2-views}
\end{figure}

During the first season, 356 plot-level video recordings were captured. For the second season, over 450 still images were taken. For each plot, the roots were laid out flat, and two orthogonal views were recorded when possible, as shown in Figure \ref{fig:orthogonal2-views}:

\begin{itemize}
    \item Top view: Capturing the flat upper side of roots.
    \item Bottom (rotated) view: Capturing the underside after rotating them.
\end{itemize}

Images and videos were captured with uniform lighting conditions and arranged to minimise shadows. In many low-severity plots (scores 1–3), only a single image was taken because damage was minimal and additional views were unlikely to provide significant benefit. Each plot’s images and videos were labelled using the expert-assigned severity score.

\subsubsection{Data Composition and Preprocessing}
\begin{enumerate}
    \item (Data Composition)
    The dataset composition and class distribution were analysed at multiple levels to understand the data structure and address imbalances. Table \ref{tab:severity-distribution1} summarises the distribution of severity scores across three stages: initial image-level, initial plot-level, and final plot-level after video frame extraction. The initial distribution (column 2) shows the images collected, which reveal a significant class imbalance. Since images were grouped by plots, with each plot containing one or two images, column 3 presents the same distribution at the plot level, better representing the experimental units. To reduce the imbalance, frames were systematically extracted from video recordings collected during the first season and incorporated into the dataset, focusing on underrepresented plot scores. The final distribution (column 4) demonstrates improved class balance, particularly for scores 5, 7, and 9. The final dataset comprised 788 images organised into 636 plots.

    \begin{table}[ht]
    \centering
    \caption{Distribution of weevil severity scores across dataset stages. Image-level counts represent individual samples, while plot-level counts correspond to grouped experimental plots. The final plot-level distribution includes frames extracted from the video to address class imbalance.}
    \label{tab:severity-distribution1}
    \begin{tabular}{|c|c|c|c|}
    \hline
    \textbf{Weevil Score} & \textbf{Initial Image-Level} & \textbf{Initial Plot-Level} & \textbf{Final Plot-Level} \\
    \hline
    1 & 215 & 109 & 109 \\
    3 & 181 & 90 & 90 \\
    5 & 28  & 12 & 289 \\
    7 & 4   & 1  & 74 \\
    9 & --  & -- & 10 \\
    \hline
    \textbf{Total} & \textbf{428} & \textbf{212} & \textbf{572} \\
    \hline
    \end{tabular}
    \end{table}

        \bigskip
    
    \textbf{Plot-level Organization}
    
    Images were organized and batched by plot (containing one or two images per plot), maintaining alignment with their corresponding expert-assigned severity scores. To further reduce bias in the dataset, additional balancing steps were applied:
    
    \begin{itemize}
        \item \textbf{Removal of severely underrepresented class:} Plots with a severity score of 9 were excluded from the study due to insufficient sample size compared to other severity classes.
        \item \textbf{Undersampling of overrepresented class:} Plots with a severity score of 5—significantly more abundant than other classes were undersampled using a stratified approach. Priority was given to plots containing two images, supplemented with single-image plots as needed, to reach a target of 110 plots (one more than the previously highest class, severity score 1, which contained 109 plots). The final distribution is shown in Table \ref{tab:final_plot_numbers}.
    \end{itemize}
    
    \begin{table}[h]
        \centering
        \caption{Final number of plots per severity score after preprocessing, per-plot batching, and class balancing.}
        \label{tab:final_plot_numbers}
        \begin{tabular}{|c|c|}
            \hline
            Severity Score&  Plots\\ \hline
            1&109\\ \hline
            3&90\\ \hline
            5&110\\ \hline
            7&74\\ \hline
        \end{tabular}
    \end{table}

    This strategy improved class balance while maintaining data quality by prioritising plots with multiple viewing angles.

    \item Preprocessing

    The preprocessing pipeline consisted of multiple stages to prepare the data for modelling. Initial analysis revealed that background elements often occupied large portions of images and exhibited visual similarities to root features, which could potentially confuse classification models. To address this, a YOLO11 segmentation model was used \citep{Khanam2024YOLOv11} and fine-tuned on a subset of manually annotated images to accurately identify and isolate roots. The segmentation model achieved a mean Average Precision (mAP) of 92.5\% on the validation set. This model was subsequently applied to all collected images to generate a clean dataset with backgrounds removed, ensuring that classification models would focus exclusively on root characteristics relevant to disease severity assessment.

    Following background removal, all images underwent uniform resizing to standard resolution for consistent input dimensions, normalisation of pixel values to ensure consistent input ranges, and Manual visual inspection to verify background removal
    
    \end{enumerate}

\subsection{Lab Assessment}
\subsubsection{Experimental Set up}

Following harvest, sweetpotato roots were thoroughly cleaned to remove soil and dirt, then organised by plot and genotype. The experimental design consisted of three replications, with each replication containing forty genotypes. A single representative root was selected from each plot to serve as the experimental unit for that genotype.

Selected roots were then transferred to the laboratory, where each root was placed in a separate, clearly labelled tin container with the corresponding genotype identifier. Containers were organised by replication to maintain experimental integrity and facilitate tracking throughout the assessment process.

\begin{figure}[h!]
    \centering
    \subfloat[]{%
      \resizebox*{8cm}{!}{\includegraphics{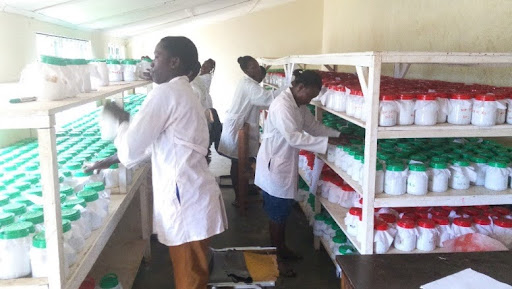}}} \hspace{10pt}
    \subfloat[]{%
      \resizebox*{5.3cm}{!}{\includegraphics{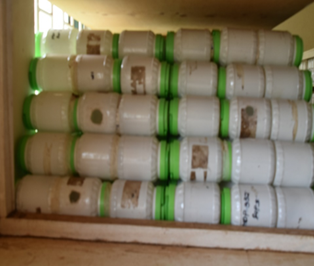}}}

    \caption{Tin containers used for the controlled infestation process. Each container held a single root and 5–10 adult weevils, sealed and incubated for 24 hours.}
    \label{fig:orthogonal-views1}
\end{figure}

Five to ten adult weevils were introduced into each labelled container, which was then sealed and allowed to incubate \citep{Mug22}. A standardised infestation protocol was implemented to ensure consistent exposure conditions across all samples. The exposure period was set at 24 hours, after which the weevils were removed and the roots sent for imaging to capture the induced damage patterns \citep{Mug22, Kye19}. Figure \ref{fig:orthogonal-views1} shows the chosen roots being placed into the containers with introduced weevils for incubation. 

After exposure, weevils were removed, and the roots were transferred for imaging to assess damage.


\subsubsection{Imaging Setup}
\begin{figure}[]
    \centering
    \includegraphics[width=1\textwidth]{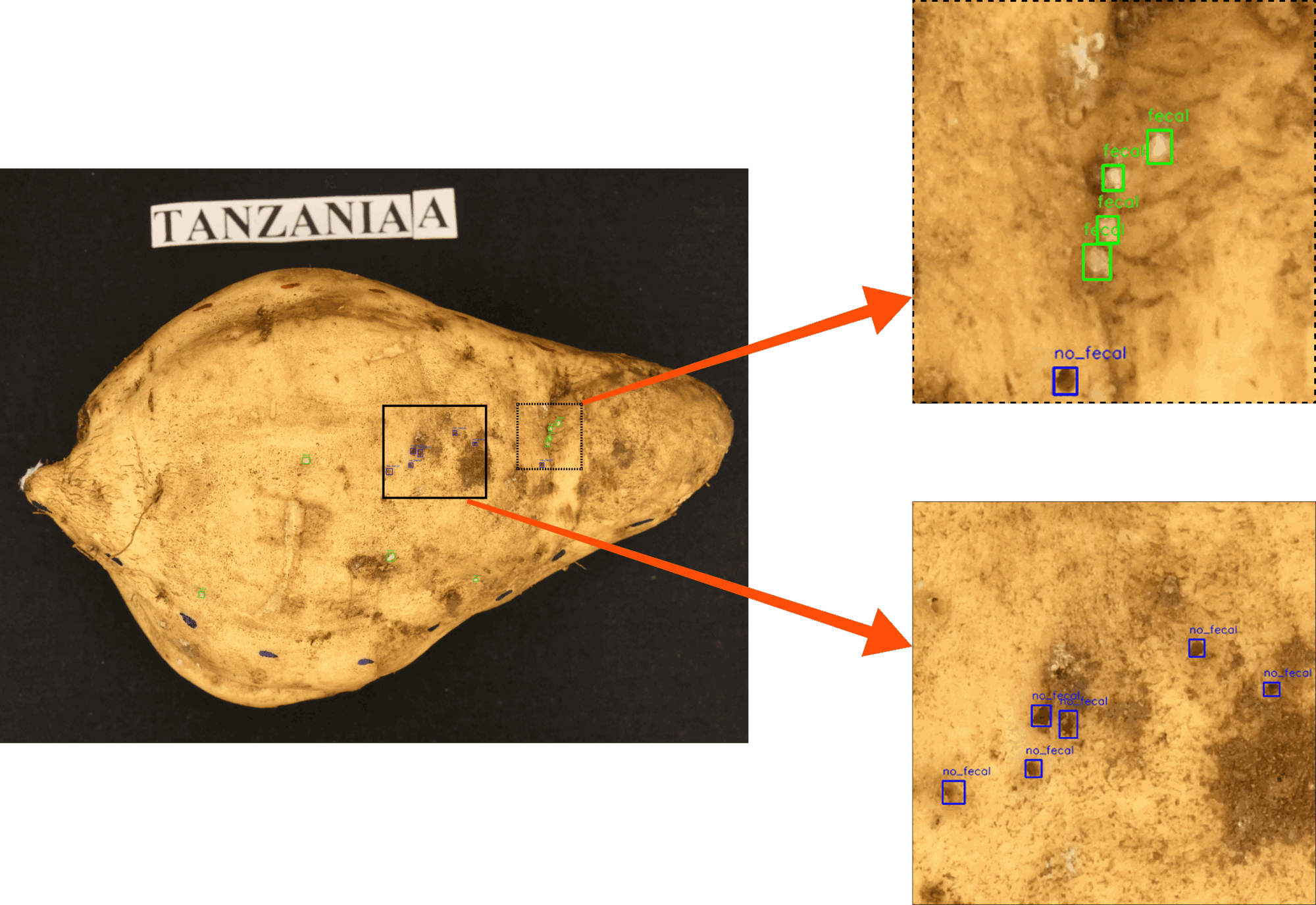}
    \caption{Sweetpotato genotype ‘Tanzania’ showing two zoomed-in views. The green boxes (top right) highlight feeding holes with fecal matter, identifiable by whitish, sponge-like tissue. These white tissues contain weevil eggs. The blue boxes (bottom right) show feeding holes without fecal matter, which appear as clean surface depressions. Accurately detecting and estimating the number of feeding holes provides an indicator of a genotype's resistance — fewer holes suggest higher resistance to sweetpotato weevil damage.}
    \label{fig:orthogonal-views}
\end{figure}

Adult weevils lay eggs on sweetpotato roots, creating distinctive whitish deposits on the surface. During feeding, weevils excavate holes that vary in appearance depending on their developmental stage and feeding behaviour. Feeding holes can be categorised into two distinct types: those containing fecal matter, characterised by white, sponge-like tissue accumulation, and those without fecal matter, which appear as clear, visible depressions or holes on the root surface, as shown in Figure \ref{fig:orthogonal-views}.

To accurately quantify weevil damage through hole counting while avoiding duplication, a systematic sectioning protocol was developed to ensure full surface coverage of each root. Roots were placed on a flat surface and divided into clearly defined sections using colored markers to eliminate overlap.

\begin{figure}[h!]
    \centering
    \subfloat[Section A]{%
      \resizebox*{6.8cm}{!}{\includegraphics{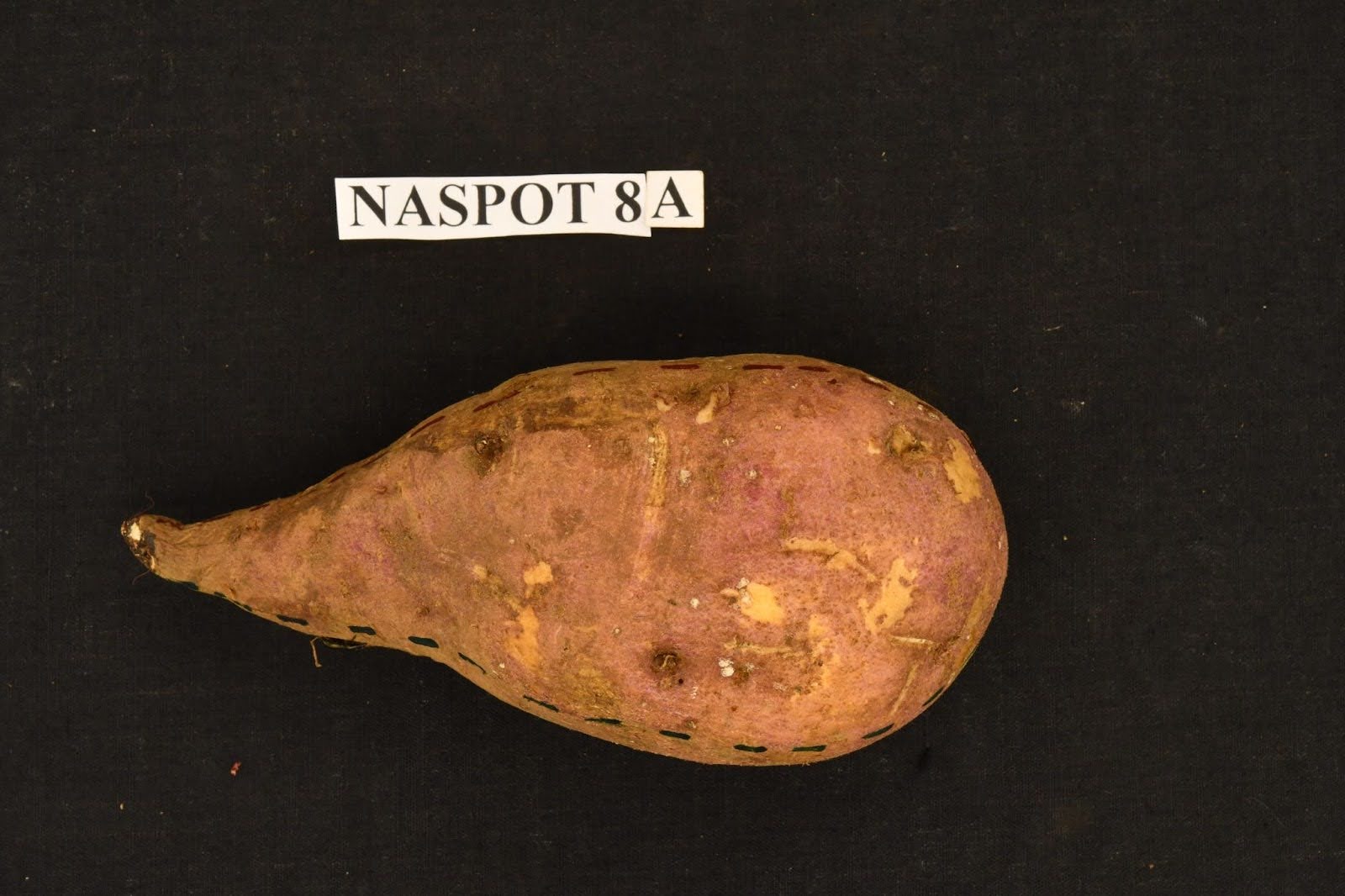}}
      \label{fig:sidea}
    }
    \hspace{8pt}
    \subfloat[Section B]{%
      \resizebox*{6.8cm}{!}{\includegraphics{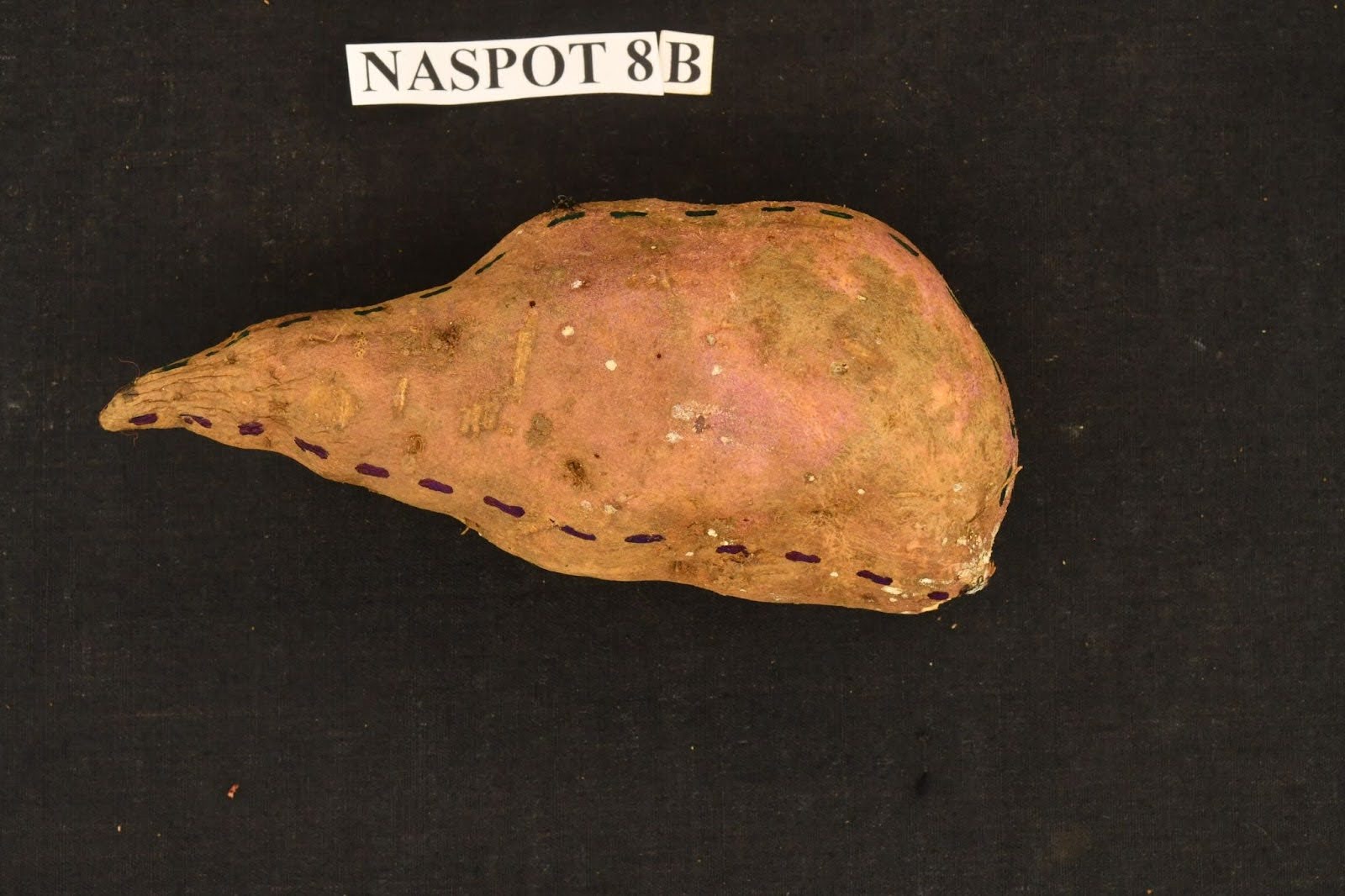}}
      \label{fig:sideb}
    }
    \hspace{8pt}
    \subfloat[Section C]{%
      \resizebox*{7cm}{!}{\includegraphics{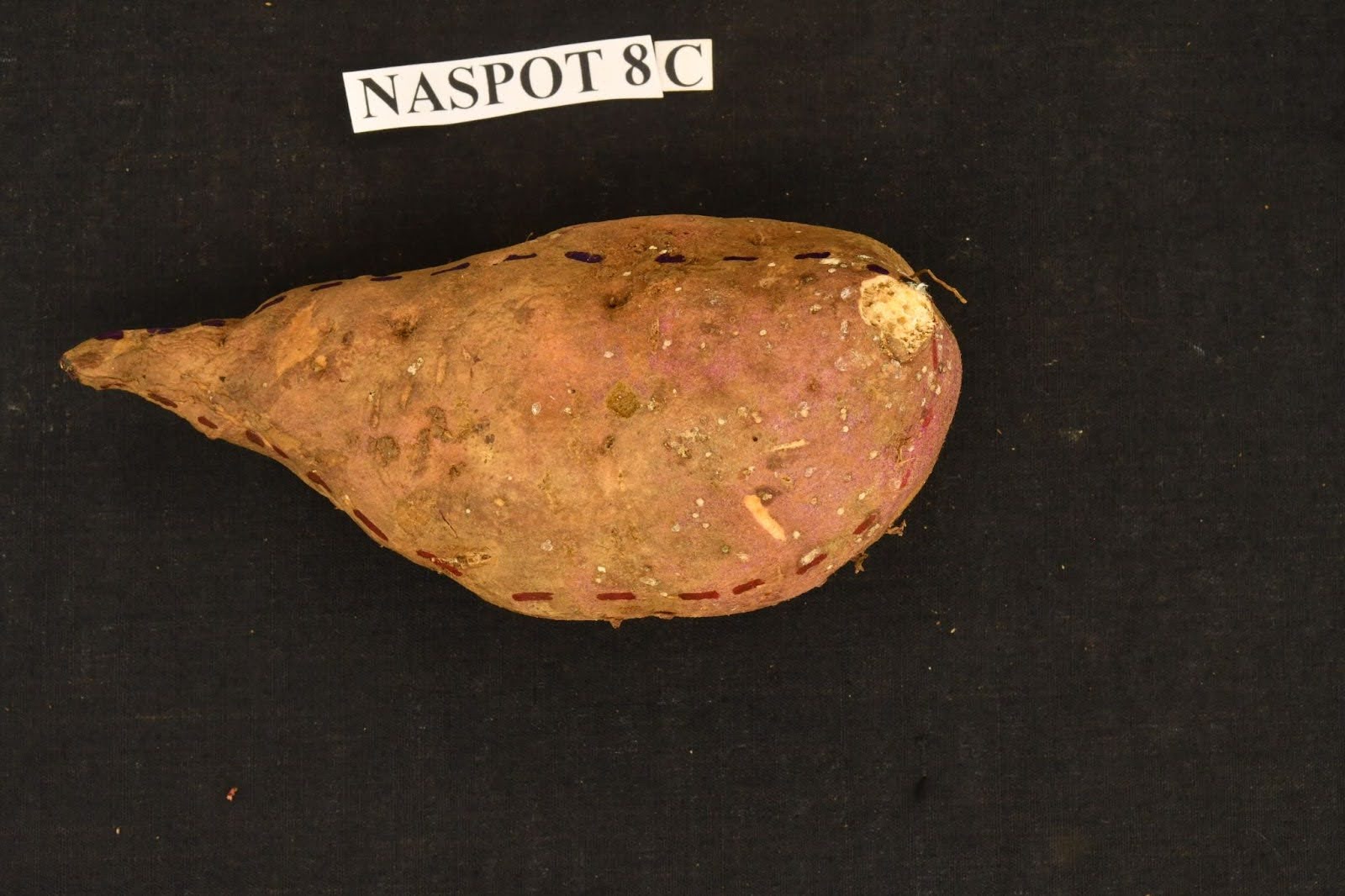}}
      \label{fig:sidec}
    }
    \caption{Three viewable sections (A, B, and C) of a single sweetpotato root after colour-coded segmentation. This approach enabled systematic imaging to avoid duplication and ensure full coverage of the sweetpotato root surface for accurate damage assessment.}
    \label{fig:roots-three-sides}
\end{figure}

The sectioning procedure was as follows: a red line marked the starting point, and a green line indicated the end of the first visible section. The region between the red and green lines formed section A, which was the focus of the first image. This pattern continued with section B (green to blue lines), section C (blue to red), and, when needed, section D (black to red). Typically, three to four sections were sufficient to capture the entire surface. Figure \ref{fig:roots-three-sides} illustrates section markings on a sample genotype.

A standardised imaging setup was used with a digital camera mounted above an Orte LED ProBox under controlled lighting. The following camera settings were used consistently across all samples:

\begin{itemize}
    \item Shutter speed: 1/250 sec
    \item Aperture: f/22
    \item Exposure compensation: 0 EV
\end{itemize}

Each captured image included a label indicating the genotype and section identifier (e.g., "NASPOT A" for the first section of genotype NASPOT). This approach yielded three to four labelled images per root, ensuring complete documentation of all surface areas for accurate damage quantification. A total of 168 images were collected.

\begin{table}[h!]
    \centering
    \caption{Summary of genotypes and total images captured per replication.}
    \label{tab:reps_count_images}
    \begin{tabular}{|c|c|c|}
    \hline
    \textbf{Replication} & \textbf{Number of genotypes} & \textbf{Total images} \\
    \hline
    1 & 40 & 121 \\
    \hline
    2 & 40 & 123 \\
    \hline
    3 & 39 & 124 \\
    \hline
    \end{tabular}
\end{table}

\subsubsection{Data Preprocessing}
\label{sec:lab-datapreprocessing}

The preprocessing pipeline was designed to transform the annotated dataset into a format optimised for small object detection, addressing both the annotation format and the challenge of detecting tiny weevil feeding holes on large root images.

The initial annotations consisted of point-based labels marking the centres of feeding holes rather than traditional bounding boxes, due to the extremely small size of sweetpotato weevil feeding holes. To convert these into a format suitable for object detection models, a bounding box estimation approach was implemented. Fixed-size bounding boxes were generated around each annotated point using an 18-pixel padding expansion, providing approximate object boundaries without requiring labour-intensive manual box drawing. Manual corrections from the experts were then applied to adjust any boxes that did not accurately capture the intended holes.

\begin{figure}[h!]
    \centering
    \subfloat[Original image showing focus area (in color), extracted by the segmentation model.]{%
      \resizebox*{6cm}{!}{\includegraphics{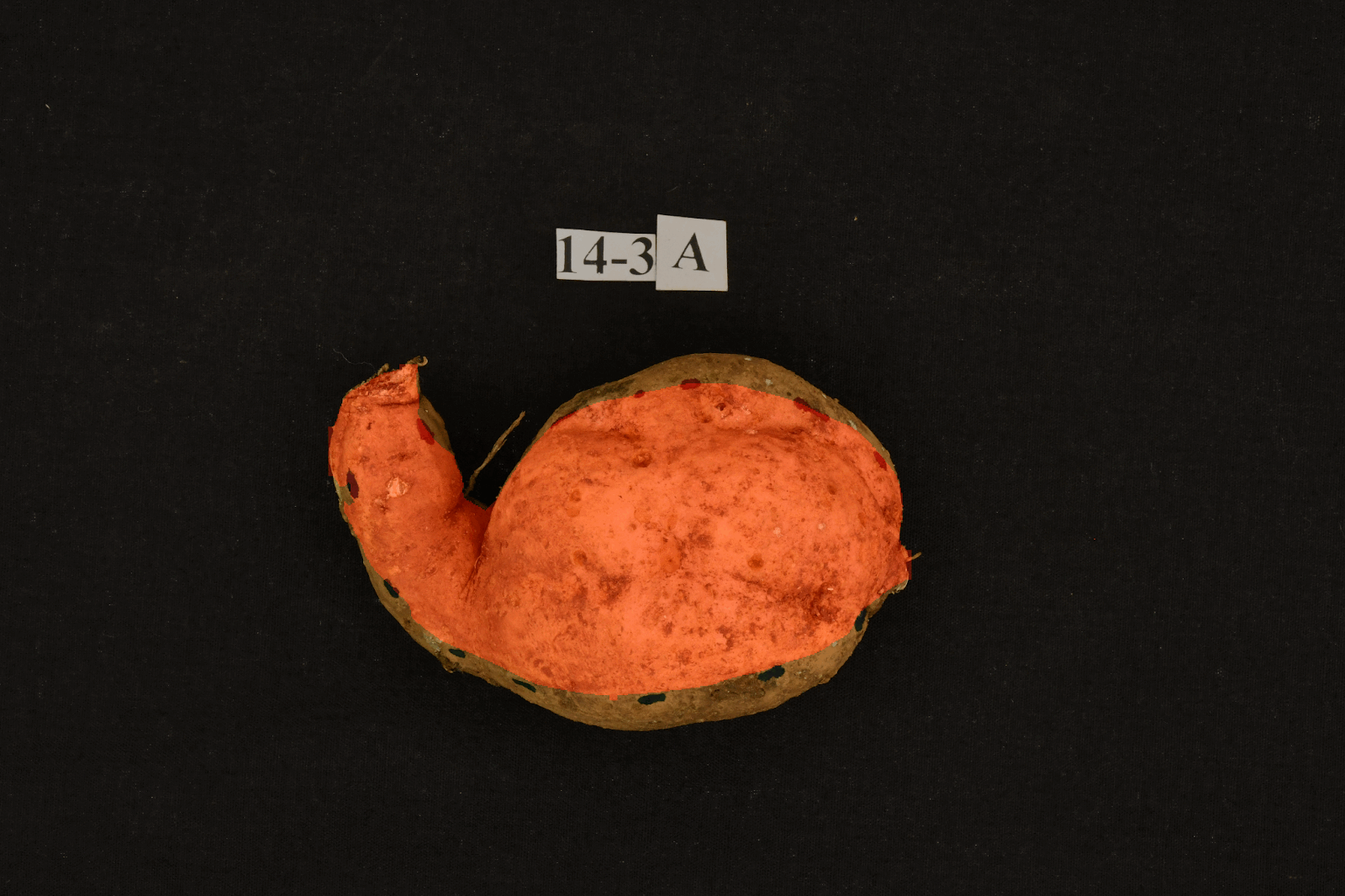}}
      \label{fig:focusarea}
    }
    \hspace{8pt}
    \subfloat[Annotated mask highlighting the root region.]{%
      \resizebox*{6cm}{!}{\includegraphics{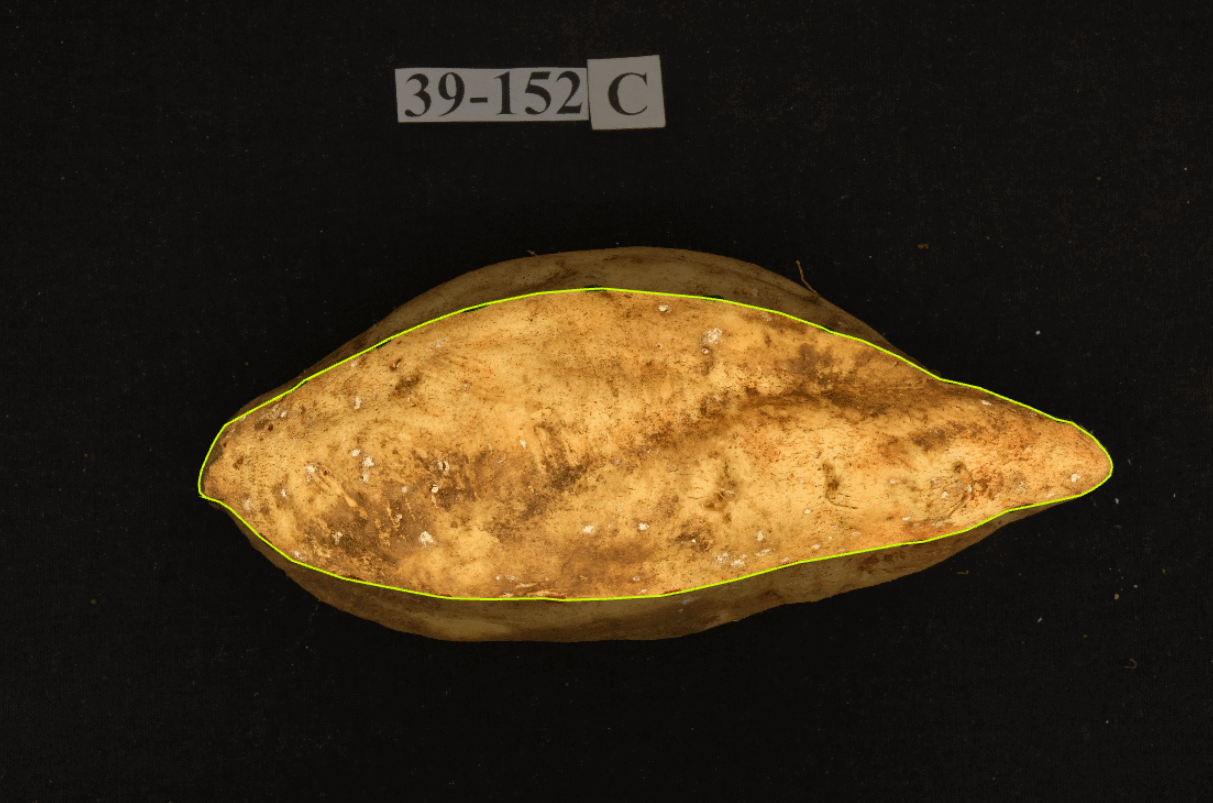}}
      \label{fig:maskimage}
    }
    \hspace{8pt}
    \subfloat[Examples of predicted masks generated by the trained YOLO11 segmentation model.]{%
      \resizebox*{12.5cm}{!}{\includegraphics{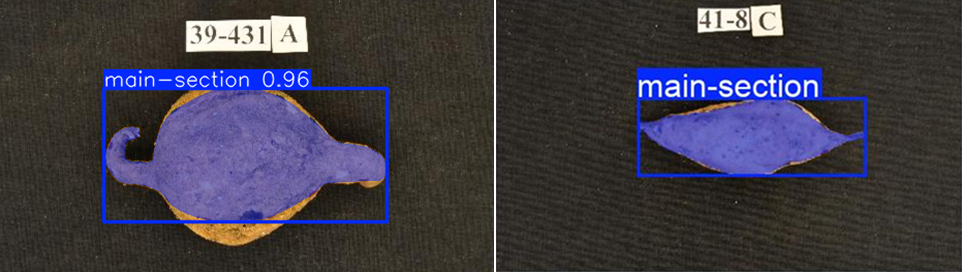}}
      \label{fig:predictedmasks}
    }
    
    \caption{Examples from the segmentation pipeline used to isolate sweetpotato root regions. a) Original image showing the detected focus area in color. b) Manually annotated mask of the root region. c) Sample predictions from the trained segmentation model.}
    \label{fig:preprocess-flow}
\end{figure}

To focus the detection model exclusively on areas susceptible to weevil damage, a segmentation approach was employed to isolate root regions from background elements. A subset of images was manually annotated to identify regions of interest, which were used to train a YOLO11 segmentation model. The model was fine-tuned on this annotated subset and achieved a mean Average Precision (mAP) of 98\% on the validation set. Figure \ref{fig:preprocess-flow} shows representative examples from this process.

\begin{figure}[h!]
    \centering
    \includegraphics[width=0.8\textwidth]{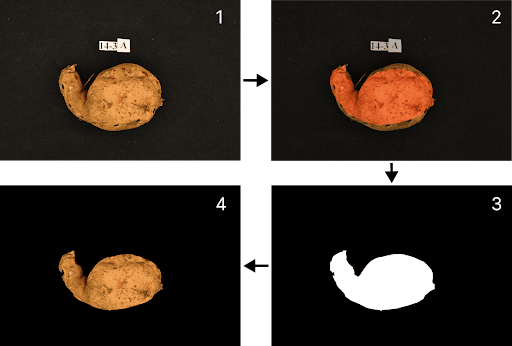}
    \caption{Preprocessing pipeline for small object detection. The figure shows four sequential stages: (1) the original annotated image, (2) the root region with the focus area mask overlaid, (3) the binary mask isolating the root area, and (4) the final segmented root region overlaid on a uniform black background.}
    \label{fig:processimage}
\end{figure}

The trained segmentation model was then applied to the full dataset to extract root regions from all images. The segmented root areas were isolated and placed against a uniform black background, effectively removing confounding background elements and creating a clean dataset focused solely on areas relevant to damage assessment, as shown in Figure \ref{fig:processimage}.

\textbf{Tiling Strategy for Small Object Detection}:
Given the extremely small size of weevil feeding holes relative to the full image dimensions, a tiling approach was implemented using the Slicing Aided Hyper Inference (SAHI) framework \citep{Aky22}. Each segmented root image was systematically divided into overlapping tiles of 384×384 pixels. This tiling strategy allowed the detection model to process high-resolution patches of the image, significantly improving the visibility and detectability of tiny feeding holes that would otherwise be overlooked in full-resolution images.

The overlapping tile configuration ensured that feeding holes located near tile boundaries would not be missed during detection. Following the tiling process, tiles containing no annotated feeding holes were removed from the dataset, as they provided no learning value for the detection model. This preprocessing pipeline resulted in a tiled dataset used for training the small object detection models.

\section{Model Development}
\subsection{Field Weevil Classification}
The dataset was split using an 80-10-10 ratio with stratification, resulting in 39 plots each for the test and validation sets, and the remaining plots were used for training. A random seed was set to ensure replicability of the results.

We implemented a dual-input ResNet-50 architecture \citep{He2016ResNet} using two identical backbones to process the paired images. We removed each model's fully connected output layer and combined the resulting 2048-d feature vectors into a 4096-d representation. This merged feature then passed through a 512-unit fully connected layer with ReLU activation, followed by a dropout layer (p = 0.5) and a final dense layer. Training was conducted for 20 epochs using a batch size of 16 with the Adam optimiser set at a learning rate of 1e-4.

We implemented a custom data loader which ensured that every sample provided the model with exactly two inputs. In cases where only one image was available, a second black image was added to maintain consistency in shape. While this approach limits the model to learning from one real view in those cases, the trade-off was essential for maintaining a simpler yet consistent input pipeline, which is crucial for deployment. We avoided heavier architectures, such as transformers, to keep the model lightweight.

A secondary model based on dual-input VGG16 \citep{Simonyan2015VGG} was similarly developed, using two shared VGG16 backbones (with partial layer freezing), followed by pooling, regularisation, and dropout before the final softmax layer. This model was trained for up to 100 epochs using SGD with momentum (learning rate = 1e-4), but it underperformed compared to the ResNet-50 version. The dual-input ResNet-50 was selected for further analysis.

The performance of the model was assessed using standard classification metrics, including accuracy, precision, recall, and the F1‑score, all derived from the confusion matrix \citep{sokolova2009systematic}. Accuracy offers a clear, overall measure of correct predictions, making it intuitive for balanced datasets. However, since our dataset was somewhat imbalanced, accuracy is not the most suitable metric for evaluating our model. Meanwhile, precision and recall provide a deeper understanding of the model’s behaviour—precision measures correctness among predicted positives, and recall measures the model’s ability to identify all actual positive cases  \citep{powers2011evaluation}. The F1‑score, as the harmonic mean of precision and recall, serves as a better single-metric summary, particularly valuable when working with imbalanced data, as in our case. We generated detailed classification reports (showing per-class precision, recall, and F1) and confusion matrices to visualise misclassification patterns and ensure comprehensive evaluation.

\subsection{Lab Weevil Detection}

Following the data preprocessing pipeline described in Section \ref{sec:lab-datapreprocessing}, the resulting tiled dataset was used to develop and train object detection models for identifying tiny weevil feeding holes. The final tiled dataset was split into training and validation sets with 3,507 images allocated for training and 500 images for validation. This split ensured sufficient training data. 
Three state-of-the-art object detection architectures were selected for comparative evaluation: YOLO11-L \citep{Khanam2024YOLOv11}, YOLO12-L \citep{Tian2025YOLOv12}, and RTDETR2 \citep{Lv2024RTDETRv2}. These models were selected based on their strong performance in object detection tasks and their ability to adapt to various scenarios upon fine-tuning. Additionally, they represent the latest and best-performing versions in their respective model families.

To ensure fair comparison across, standardised training configurations were applied to all models:
The model input size was matched to the tile dimensions (384 * 384) to eliminate unnecessary resizing operations and preserve resolution. All models were trained for 500 epochs with early stopping using a patience of 30 epochs to prevent overfitting. A batch size of 16 was used. Confidence thresholds were set to 0.6 for classification (cls), 0.5 for bounding box regression (box), and 0.5 for Intersection over Union (IoU). The Adam optimiser was employed for YOLO11 and YOLO12 models with an initial learning rate (lr0) of 0.0005 and a final learning rate (lrf) of 0.001. RTDETR2 utilised its default optimisation scheme due to architectural differences. Given the varying sizes of feeding holes within the dataset, multi-scale training was enabled.

Model performance was also evaluated using standard object detection evaluation metrics to assess detection accuracy and reliability. Mean Average Precision (mAP) served as the primary evaluation metric, calculated across different Intersection over Union (IoU) thresholds to assess both localisation and classification performance \citep{everingham2010pascal, lin2014microsoft}. Additionally, precision, recall, and F1-score were also computed to provide information on model behaviour. Precision quantifies the proportion of correct positive predictions among all positive predictions, while recall measures the proportion of actual feeding holes successfully detected by the model, and the F1 score provides a harmonic mean of precision and recall \citep{powers2011evaluation}.

\section{Results}
\subsection{Field Weevil Classification}
The best-performing classification model achieved a test accuracy of 71.43\% on unseen data, demonstrating reasonable capability in distinguishing between different weevil damage severity levels. However, the validation accuracy of 60.00\% revealed a notable performance gap, suggesting challenges in generalising across the validation dataset. This discrepancy, coupled with a validation loss of 0.85, indicates that the model's predictions lacked consistent confidence.

Table \ref{tab:classification-report} presents detailed per-class performance metrics across the four severity categories evaluated. The classification results revealed varying performance across different damage severity levels, with distinct patterns emerging for each class.

\begin{table}[h]
    \centering
    \caption{Per-class classification performance on the test set, including precision, recall, F1‑score, and support for each weevil damage severity level. The metrics highlight the model's varying ability to correctly classify severity classes 1, 3, 5, and 7.}
    \label{tab:classification-report}
    \begin{tabular}{c c c c r}
    \hline
    \textbf{Class} & \textbf{Precision} & \textbf{Recall} & \textbf{F1‑score} & \textbf{Support} \\
    \hline
    1 & 0.64 & 0.82 & 0.72 & 11 \\
    3 & 0.71 & 0.56 & 0.62 & 9 \\
    5 & 0.71 & 0.45 & 0.56 & 11 \\
    7 & 0.55 & 0.75 & 0.63 & 8 \\
    \hline
    \multicolumn{4}{r}{\textbf{Total}} & \textbf{39} \\
    \hline
    \end{tabular}
\end{table}

Severity class 1 (minimal damage) demonstrated the most balanced performance, achieving 0.64 precision and 0.82 recall, resulting in the highest F1-score of 0.72. This strong recall performance indicates the model's effectiveness in identifying plots with minimal weevil damage, which is crucial for selecting resistant genotypes.
On the other hand, severity class 3 (moderate damage) exhibited moderate performance, with a precision of 0.71 but a lower recall of 0.56, suggesting that the model tends to be conservative in predicting this intermediate damage level. 
Severity class 5 (severe damage) exhibited the most challenging classification performance, with a precision of 0.71 but the lowest recall of 0.45. This pattern indicates frequent misclassification of severe damage cases, potentially confusing them with other severity levels. Severity class 7 (very severe damage) demonstrated high recall (0.75) but lower precision (0.55), suggesting the model tends to over-predict this extreme damage category. Despite this imbalance, the F1-score of 0.63 indicates reasonable overall performance for detecting the most severely damaged plots.

\begin{figure}[H]
    \centering
    \includegraphics[width=0.5\textwidth]{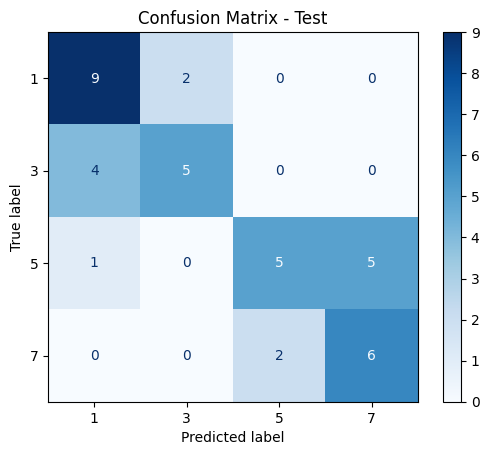}
    \caption{Confusion matrix showing the classification results across weevil damage severity levels. Most misclassifications occur between adjacent classes (e.g., class 1 vs. class 3, and class 5 vs. class 7), highlighting the model's difficulty in distinguishing between neighbouring severity categories.}
    \label{fig:confusion-matrix}
\end{figure}

Figure \ref{fig:confusion-matrix} illustrates the model’s classification outcomes using a confusion matrix. Notably, there is observable confusion between severity classes 1 and 3, as well as between classes 5 and 7. The same can be said for class 5 and class 7. These misclassification patterns suggest that adjacent severity levels present the greatest challenge for the model, likely due to overlapping visual characteristics

\subsection{Lab Weevil Detection}

Three state-of-the-art object detection models, YOLO11, YOLO12, and RT-DETR2, were trained and evaluated for the automated detection of weevil feeding holes in sweetpotato roots. All models were trained on the tiled dataset of 384×384 pixel images, with performance monitored through loss functions and evaluation metrics throughout the training process.

Training convergence was achieved across all models, with box loss, classification loss (cls-loss), and distribution focal loss (dfl-loss) showing steady decline across both training and validation sets, indicating consistent learning without overfitting. Precision and recall metrics demonstrated stable improvement over training epochs for all model architectures.

\renewcommand{\arraystretch}{1.5}
\setlength{\tabcolsep}{10pt}

\begin{table}[h]
    \centering
    \caption{Per-class evaluation of YOLO11, YOLO12, and RT-DETR2 models using precision, recall, and mAP@0.5 metrics for feeding holes with fecal matter (“Fecal”) and those without fecal matter (“No Fecal”) detection.}
    \label{tab:version-class-metrics}
    \begin{tabular}{|l|c|c|c|c|c|c|c|c|c|}
    \hline
    \textbf{Version} & \textbf{Class} & \textbf{Precision} & \textbf{Recall} & \textbf{mAP@0.5} \\
    \hline
    YOLO11 & All       & 0.707 & 0.661 & 0.685 \\
            & Fecal     & 0.692 & 0.665 & 0.69 \\
            & No Fecal  & 0.722 & 0.658 & 0.68 \\
    \hline
    YOLO12 & All       & 0.715 & 0.765 & \textbf{0.777} \\
            & Fecal     & 0.726 & 0.714 & 0.728 \\
            & No Fecal  & 0.705 & 0.815 & 0.826 \\
    \hline
    RT-DETR2 & All       & 0.746 & 0.622 & 0.69 \\
            & Fecal     & 0.752 & 0.634 & 0.693 \\
            & No Fecal  & 0.74 & 0.611 & 0.688 \\
    \hline
    \end{tabular}
\end{table}

Table \ref{tab:version-class-metrics} presents the evaluation results for all three models across both feeding hole categories. YOLO12 emerged as the best-performing model, achieving the highest overall mAP@0.5 score of 0.777, followed by RT-DETR2 (0.69) and YOLO11 (0.685). The superior performance of YOLO12 was particularly evident in its balanced precision-recall trade-off, with a precision of 0.715 and a recall of 0.765 across all classes.

Analysing class-specific performance, all models demonstrated consistent detection capabilities across both faecal and non-faecal feeding hole categories. YOLO12 showed the most robust performance for detecting holes with fecal matter (mAP@0.5: 0.728), while achieving exceptional performance on non-fecal holes (mAP@0.5: 0.826). RT-DETR2 exhibited the highest precision (0.746) but suffered from lower recall (0.622), suggesting a more conservative detection approach that may miss some positive instances. A probable reason for YOLO12’s superior performance is its attention‑centric real‑time architecture.

\begin{figure}[h!]
    \centering
    \includegraphics[width=\textwidth]{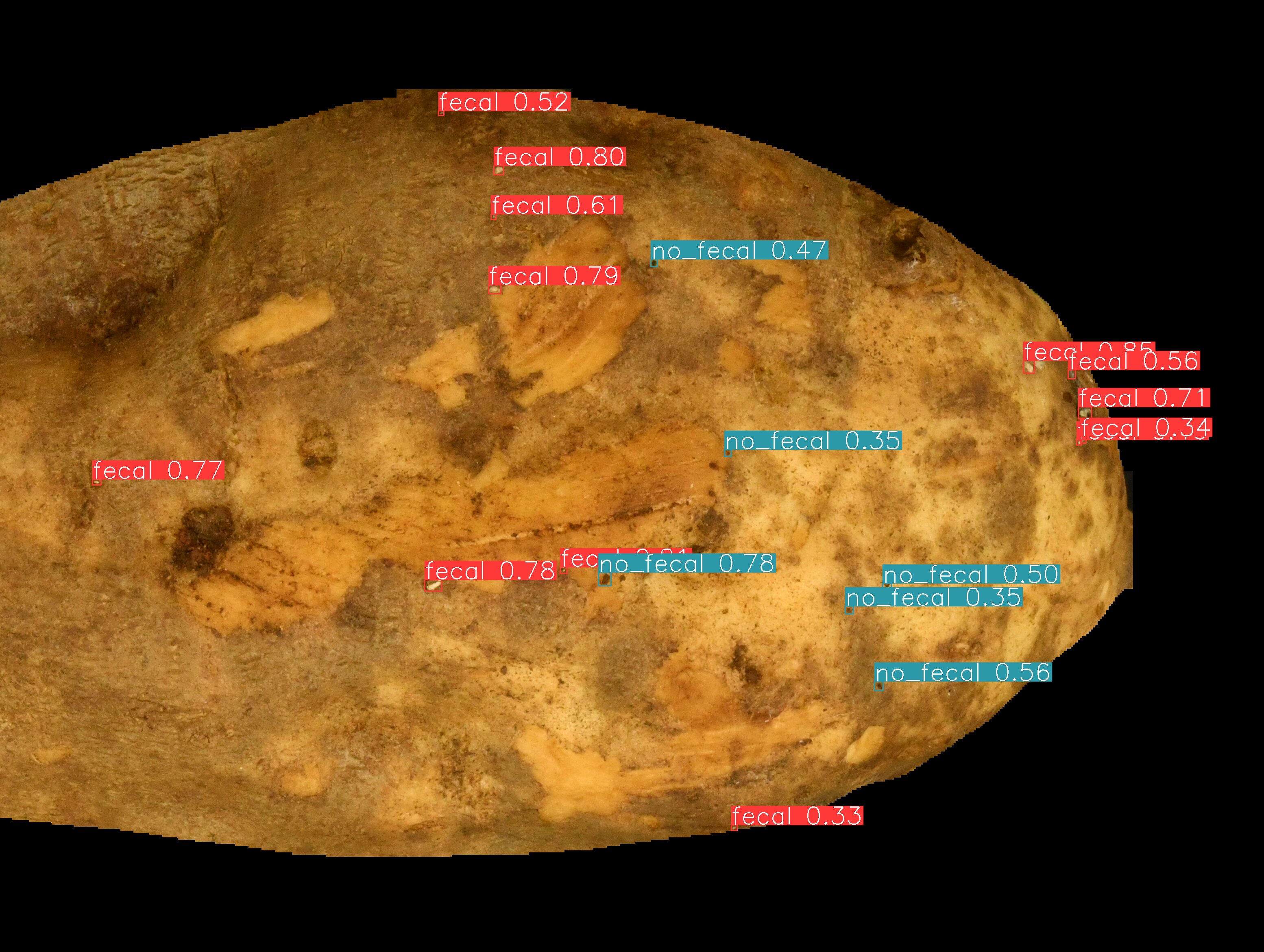} \\
    \includegraphics[width=\textwidth]{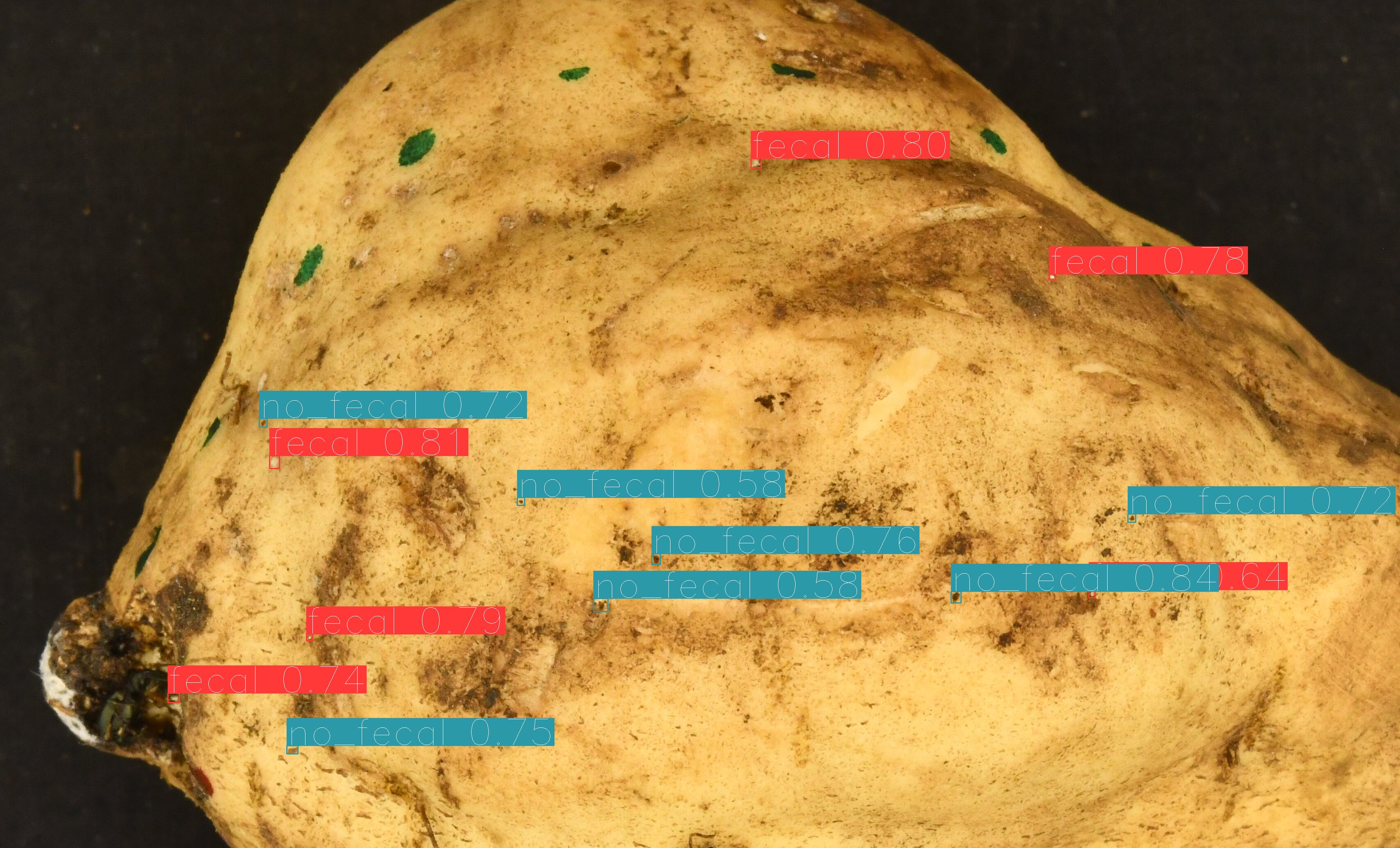}

    \caption{Sample outputs from the YOLO12 model showing predicted bounding boxes for weevil feeding holes on zoomed-in sweetpotato roots. Both fecal and non-fecal holes are detected. The feeding holes are extremely small, making them difficult to visualize clearly in full-resolution images without zooming in.}
    \label{fig:yolo12-examples}
\end{figure}

To qualitatively validate the detection results, representative output images from the best-performing model (YOLO12) are presented in Figure \ref{fig:yolo12-examples}. These examples illustrate the model’s ability to accurately localize both fecal and non-fecal feeding holes, even when they appear in clusters or near root edges.

\bigskip

Based on the evaluation, YOLO12 was selected as the optimal model for weevil feeding hole detection, demonstrating the best balance of precision, recall, and overall detection accuracy. The model's superior performance on both damage categories, combined with its robust training convergence, makes it most suitable for practical deployment in sweetpotato weevil assessment applications.

\section{Discussion}
The results demonstrate the feasibility of automating sweetpotato weevil damage assessment using computer vision, although performance varied depending on the context. Models trained on lab data outperformed those trained on field data, with YOLO12 achieving a 77.7\% mAP@0.5 in controlled conditions, compared to a 71.43\% test accuracy for the field classification model. This disparity reflects the challenges of field environments, such as inconsistent lighting, varied root appearances, and subjective scoring, versus the standardised imaging, uniform backgrounds, and controlled weevil exposure in the lab. Notably, field classification showed higher recall for extreme severity classes (1 and 7), likely due to their more visually distinct features.

These findings have significant implications for sweetpotato breeding programs. The field model's strong performance in identifying minimal damage cases (82\% recall for class 1) is especially useful for flagging resistant genotypes during early screening. However, lower accuracy in intermediate classes suggests automated systems are better suited as decision-support tools than full replacements for expert assessment. In contrast, lab-based detection enables objective quantification of damage through feeding hole counts, helping to standardise evaluations and reduce subjective bias. Together, these approaches can offer a promising pipeline in combination: field-level screening followed by lab-based verification for breeding selection.

Despite encouraging results, several technical limitations must be acknowledged. The large gap between test and validation accuracy in the field classifier suggests overfitting, likely due to limited training data and high variability in field conditions. Misclassifications were most frequent between adjacent severity levels, highlighting the need for more objective, standardised field scoring protocols and data. Additionally, the two-stage lab detection pipeline, while accurate, introduces computational overhead that may limit real-time use in the field.

From a deployment perspective, lightweight object detection models like YOLO11 and YOLO12 show potential for integration into edge devices, though further optimisation for smartphones or tablets is necessary. Techniques such as model pruning, quantisation, or architectural refinement could reduce computational demands without sacrificing performance. These methods align well with existing breeding workflows and have the potential to reduce evaluation time while improving consistency. As models continue to improve and hardware access expands, pilot deployment in agricultural research settings appears both feasible and beneficial.

\section{Conclusion and Future Work}
This study demonstrates the viability of using computer vision and deep learning for automated assessment of sweetpotato weevils in both field and laboratory settings. The models achieved encouraging performance—YOLO12 reached a mAP@0.5 of 77.7\% for detecting feeding holes in lab images, while the field-based classifier achieved 71.43\% test accuracy in predicting severity levels. These results represent a significant step toward developing scalable, objective phenotyping systems that can complement traditional manual evaluations, which are often time-consuming and prone to subjectivity.

The methods introduced—such as the two-stage lab detection pipeline with segmentation and tiling—effectively tackle the challenge of identifying small-scale feeding damage, while the field model supports rapid screening of plot-level severity. Though not yet a replacement for expert judgment, these tools show considerable promise for improving the speed, consistency, and efficiency of breeding workflows. Importantly, the datasets and code will be made publicly available to support further research and model development within the plant phenotyping community.

Looking ahead, efforts will focus on building a unified, user-friendly platform that integrates both models into a single interface. This would allow breeding programs and researchers to upload root images and receive automated evaluations—damage counts in lab settings and severity estimates in the field—potentially through a mobile app to support in-field usage.

Future technical work should aim to enhance field classification performance through more expressive model architectures, such as those using attention mechanisms, transformers, or multi-scale feature aggregation. Increasing dataset diversity across seasons and locations will also help improve generalisation under varied field conditions. Additionally, incorporating multimodal inputs—such as spectral or thermal imaging—could yield more informative assessments and a deeper understanding of plant-pest interactions.

\section*{Acknowledgement(s)}

The authors gratefully acknowledge the National Agricultural Research Organisation (NARO) of Uganda and the International Potato Center (CIP) for their collaboration and support throughout this research.

\bigskip
\bigskip

\end{document}